\documentclass[10pt,twocolumn]{article}

\usepackage{cvpr}
\usepackage{times}
\usepackage{epsfig}
\usepackage{graphicx}
\usepackage{amsmath}
\usepackage{amssymb}

\usepackage{multirow}
\usepackage{footmisc}
\usepackage{color}
\usepackage{threeparttable}
\usepackage{soul}
\usepackage{nopageno}

\usepackage{todo}

\usepackage[pagebackref=true,breaklinks=true,colorlinks,bookmarks=false]{hyperref}

\cvprfinalcopy % *** Uncomment this line for the final submission

\def\cvprPaperID{5581} % *** Enter the CVPR Paper ID here

\ifcvprfinal\fi

\begin{document}

%%%%%%%%% TITLE
\title{Can 3D Adversarial Logos Cloak Humans?}

\author{Yi Wang\textsuperscript{1}\footnotemark[1], Jingyang Zhou\textsuperscript{2}\footnotemark[1] , Tianlong Chen\textsuperscript{1}, Sijia Liu\textsuperscript{3}, Shiyu Chang\textsuperscript{3}, Chandrajit Bajaj\textsuperscript{1}, Zhangyang Wang\textsuperscript{1}\\
\textsuperscript{1} University of Texas at Austin, \\ \textsuperscript{2}University of Science and Technology of China, \textsuperscript{3}MIT-IBM Watson AI Lab \\
\textit{\small \{panzer.wy, tianlong.chen, bajaj, atlaswang\}@utexas.edu}, \\
\textit{\small djycn1996@mail.ustc.edu.cn}, 
\textit{\small \{sijia.liu, shiyu.chang, lisa.amini\}@ibm.com},
% \url{https://github.com/TAMU-VITA/3D_Adversarial_Logo}
}

\maketitle
%\thispagestyle{empty}

%%%%%%%%% ABSTRACT
\begin{abstract}
% This paper presents a new adversarial logo attack from 3D rendering to make human invisible in front of cameras: we construct a target logo from a 2D texture image and map this image into a 3D adversarial logo via a texture mapping called logo transformation. The resulting 3D adversarial logo is then viewed as an adversarial texture enabling easy manipulation of its shape and position. This greatly extends the versatility of adversarial training. Contrary to the traditional patch-based adversarial attacks, where prior work attempts to fool trained object detectors using appended adversarial patches, this new form of attack is mapped into the 3D object world and back-propagates to the 2D image domain through differentiable rendering. In addition, and unlike existing adversarial patches, our new 3D adversarial logo is shown to fool state-of-the-art deep object detectors robustly under varying camera views, leading to a potential success that our scheme is persistently strong in the physical world.
With the trend of adversarial attacks, researchers attempt to fool trained object detectors in 2D scenes. Among many of them, an intriguing new form of attack with potential real-world usage is to append adversarial patches (\eg logos) to images. Nevertheless, much less have we known about adversarial attacks from 3D rendering views, which is essential for the attack to be persistently strong in the physical world. This paper presents a new 3D adversarial logo attack: we construct an arbitrary shape logo from a 2D texture image and map this image into a 3D adversarial logo via a texture mapping called logo transformation. The resulting 3D adversarial logo is then viewed as an adversarial texture enabling easy manipulation of its shape and position. This greatly extends the versatility of adversarial training for computer graphics synthesized imagery. Contrary to the traditional adversarial patch, this new form of attack is mapped into the 3D object world and back-propagates to the 2D image domain through differentiable rendering. In addition, and unlike existing adversarial patches, our new 3D adversarial logo is shown to fool state-of-the-art deep object detectors robustly under model rotations, leading to one step further for realistic attacks in the physical world. Our codes are available at \url{https://github.com/TAMU-VITA/3D_Adversarial_Logo}.

%and rendered back to 2D scenarios as part of 3D human models. When we feed the images (with humans and backgrounds) synthesized from different views into state-of-the-art object detectors, the change of loss function back-propagates to the adversarial logo defined on the 2D domain through differentiable rendering, while every 3D human mesh changes its texture concurrently based on the mapping we predefined. 
%We extend our work to multi-mesh joint training that can effectively attack different (even unseen) meshes. 

\end{abstract}
\renewcommand{\thefootnote}{\fnsymbol{footnote}}
\footnotetext[1]{Equal Contribution.}

%%%%%%%%% BODY TEXT
% Every section is put in an individual file.
\section{Introduction}

\begin{figure}[t]
\centering
\includegraphics[width=1\linewidth]{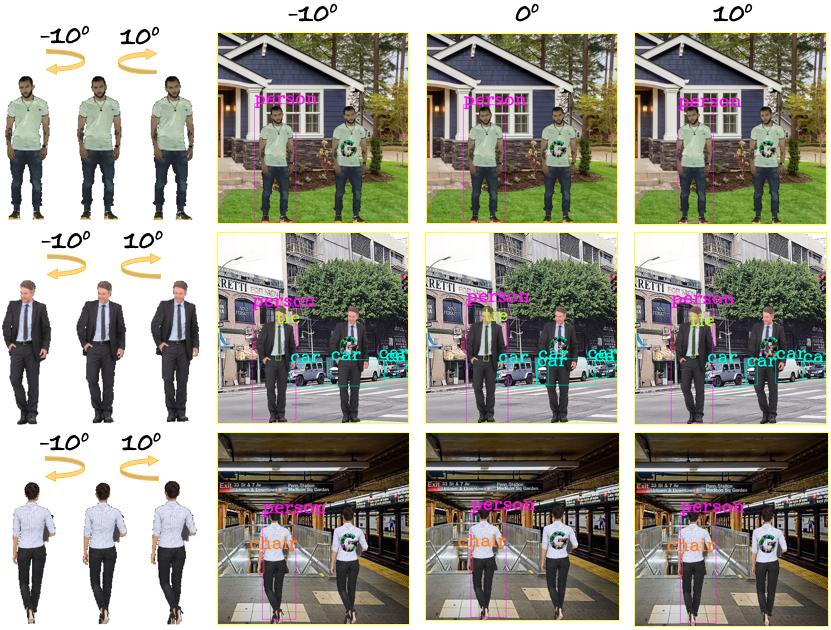}
\caption{Examples of our 3D adversarial logo attack on different 3D object meshes to fool a YOLOV2 detector. The 3D adversarial patch (as a logo ``G") is viewed as part of the textures map over 3D human mesh models. When rendering the 3D mesh scene with implanted 3D adversarial logos, and from multiple different angle views (from $-10$ to $10$ degrees) and with human postures, the attack stays robust causing mis-recognition, i.e. making recognized humans ``disappear". 
The first, second and third column shows rendering results for a -10 degree angle view, a 0 degree angle view and a 10 degree angle view, respectively. In each case, the human with our adversarial logo is not recognized by a YOLOV2 detector.}
% \vspace{-5mm}
\label{fig:intro}
\end{figure}

Deep neural networks are notoriously vulnerable: to human-imperceivable perturbations or doctoring of images, resulting in the trained algorithms, drastically changing their recognition and predictions. To test the mis-recognition or mis-detection vulnerability, \cite{tramer2017ensemble} propose 2D adversarial attacks, manipulating pixels on the image while maintaining overall visual fidelity. This negligible perturbation to human eyes causes drastically false conclusions with high confidence by trained deep neural networks. Numerous adversarial attacks have been designed and tested on deep learning tasks such as image classification and object detection. Among extensive efforts, the focus recently has shifted to only structurally editing certain local areas on an image, known as \textit{patch adversarial attacks} \cite{brown2017adversarial}. Thys et al.\cite{thys2019fooling} propose a pipeline to generate a 2D adversarial patch and attach it to image pixels of humans appearing in 2D images. In principle, a person with this 2D adversarial patch will fool or become ``invisible'' from deep learned human image detectors. However, such 2D image adversarial patches are often not robust to image transformations, and especially under multi-view  2D image synthesis in reconstructed 3D computer graphics settings. Examining 2D image renderings from 3D scene models using various possible human postures and different angle-view of humans, the 2D attack can easily lose its own strength under such 3D viewing transformations. 
Moreover, while square or rectangular adversarial patches are typically under consideration, more shape variations and their implications for the attack performance have rarely been discussed before. 

Can we naturally stitch a patch onto human clothes to make the adversarial attack more versatile and realistic? The defect in pure 2D scenarios leads us to consider the 3D adversarial attack, where we view a person as a 3D object instead of its 2D projection. As an example, the domain of mesh adversarial attack \cite{xiao2019meshadv} refers to deformations in the mesh's shape and texture level to fulfill the attack goal. However, these 3D adversarial attacks were not yet justified the concept of patch adversarial attack; they view the entire texture and geometric information of 3D meshes as attackable. Moreover, a noticeable branch of researches shows that 2D images with infinitesimal rotation and shift may cause huge perturbation in predictions  \cite{zhang2019making,azulay2018deep,engstrom2019exploring}, no matter how negligible to human eyes. What if the perturbation does not come from 2D scenarios and conditions ($e.g.$, 2D rotation and translation), but rather results from physical world change, like 3D view rotations and body postures changes? Furthermore, effective attacks on certain meshes do not imply a generalized effectiveness among other meshes, $e.g.$, the attack can be failed when changing to a different clothes mesh. Those downsides motivate us to develop more generalizable 3D adversarial patches. 

The primary aim of this work is to generate a structured patch in an arbitrary shape (called a ``logo'' by us), termed as a \textit{3D adversarial logo} that, when appended to a 3D human mesh, then rendered into 2D images, can consistently fool the object detector under different human postures %and from different camera views
. A 3D adversarial logo is defined over a subregion on 3D mesh that can alter the textures and position. Then 3D human meshes along with 3D adversarial logos are rendered on top of real-life background images. The specific contributions of our work are highlighted as: 
\begin{itemize}
\item [$\bullet$] We propose a logo transformation pipeline to map an arbitrary 2D shape (``logo") into a mesh to form the proposed 3D adversarial logos. Moreover, the 3D adversarial logo is updated when the loss is propagated back from the 2D adversarial logo, and eventually propagated to the texture image. The pipeline can be easily extended to multiple-mesh joint training. 

\item [$\bullet$] We propose a general 3D-to-2D adversarial attack protocol via physical rendering equipped with differentiability. We render 3D meshes, with the 3D adversarial logo attached on, into 2D scenarios and synthesize images that could fool the detector. The shape of our 3D adversarial logo comes from the selected logo texture in the 2D domain. Hence, we can perform versatile adversarial training with shape and position controlled.

\item [$\bullet$] We justify that our model can adapt to multi-angle scenarios with much richer variations than what can be depicted by 2D perturbations, taking one important step towards studying the physical world fragility of deep networks. 

\end{itemize}

\section{Related Work}
\subsection{Differentiable Mesh}
Various tasks, including depth estimation as well as 3D reconstruction from 2D images, have been explored with deep neural networks and witnessed successes. Less considered is the reverse problem: How can we render the 3D model back to 2D images to fulfill desired tasks? 

Discrete operations in the two most popular rendering methods (ray-tracing and rasterization) hamper the differentiability. To fill in the gap, numerous approaches have been proposed to edit mesh texture via gradient descent, which provides the ground to combine traditional graphical renderer with neural networks. Nguyen-Phuoc et al. \cite{nguyen2018rendernet} propose a CNN architecture leveraging a projection unit to render a voxel-based 3D object into 2D images. Unlike the voxel-based method, Kato et al. \cite{kato2018neural} adopt linear-gradient interpolation to overcome vanishing gradients in rasterization-based rendering. Raj et  al. \cite{raj2019learning} generate textures for 3D mesh through photo-realistic pictures. They then apply RenderForCNN \cite{su2015render} to sample the viewpoints that match the ones of input images, followed by adapting CycleGAN \cite{zhu2017unpaired} to generate textures for 2.5D information rendered in the generated multi-viewpoints, and eventually merge these textures into a single texture to render the object into the 2D world.

\subsection{Adversarial Patch in 2D Images}
Adversarial attacks \cite{szegedy2013intriguing,goodfellow2014explaining,tkhu2019triplewins,Chen_2020_CVPR,gui2019ATMC} are proposed to analyze the robustness of CNNs, and recently are increasingly studied in object detection tasks, in the form of adversarial patches. For example, \cite{chen2018shapeshifter} provides a stop sign attack to Fast-RCNN \cite{girshick2015fast}, and \cite{thys2019fooling} is fooling the YOLOv2 \cite{redmon2017yolo9000} object detector through pixel-wise patch optimization. The target patch with simple 2D transformations (such as rotation and scaling) is applied to a near-human region in 2D real photos and then trained to fool with the object detector. To demonstrate realistic adversarial attacks, they physically let a person hold the 3D-printed patch and verify them to "disappear" in the object detector. Nevertheless, such attacks are easily broken w.r.t. real-world 3D variations as pointed out by \cite{lu2017no}. Wiyatno et al.  \cite{wiyatno2019physical} propose to generate physical adversarial texture as a patch in backgrounds. Their method allows the patch "rotated" in 3D space and then added back to 2D space. Xu et al.  \cite{xu2019evading} discusses how to incorporate physical deformation of T-shirts into patch adversarial attacks, leading a forward step yet only in a fixed camera view.  A recent work by Huang et al. \cite{huang2020universal} attacks region proposal network (RPN) by synthesizing semantic patches that naturally anchored onto human cloth in the digital space. They test the garment in the physical world with motions and justify their result in digital space.

\subsection{Mesh Adversarial Attack}
A 2D object can be considered as a projection from the 3D model. Therefore, attacking from 3D space and then map to 2D space can be seen as a way of augmenting perturbation space. %Anish et al. \cite{athalye2017synthesizing} attests such possibility to create 3D adversarial objects in the physical world. 
In recent two years, different adversarial attacks scheme over 3D meshes have been proposed. For instance, Tsai et al. \cite{tsai2020robust} perturb the position of the point cloud to generate an adversarial mesh that fools 3D shape classifiers. Ti et al. \cite{liu2018beyond} alter lighting and geometry information of a physical model, to generate adversarial attacks, by modeling the pixel in natural images as an interaction result of lighting condition and physical scene, such that the pixel can maintain its natural appearance. More recently, Xiao et al. \cite{xiao2019meshadv} and Zeng et al. \cite{zeng2019adversarial} generate adversarial samples by altering the physical parameters ($e.g. $ illumination) of rendering results from target objects. They generate meshes with negligible perturbations to the texture and show that under certain rendering assumptions($e.g.$ fixed camera view), the adversarial mesh can remain to deceive state-of-the-art classifiers and detectors. Overall, most existing works perturb the image global texture $etc.$, while the idea of generating an adversarial sub-region/patch remains unexplored in the 3D mesh domain.

\section{The Proposed Framework}

\begin{figure*}[t]
\centering
\includegraphics[width=1\linewidth]{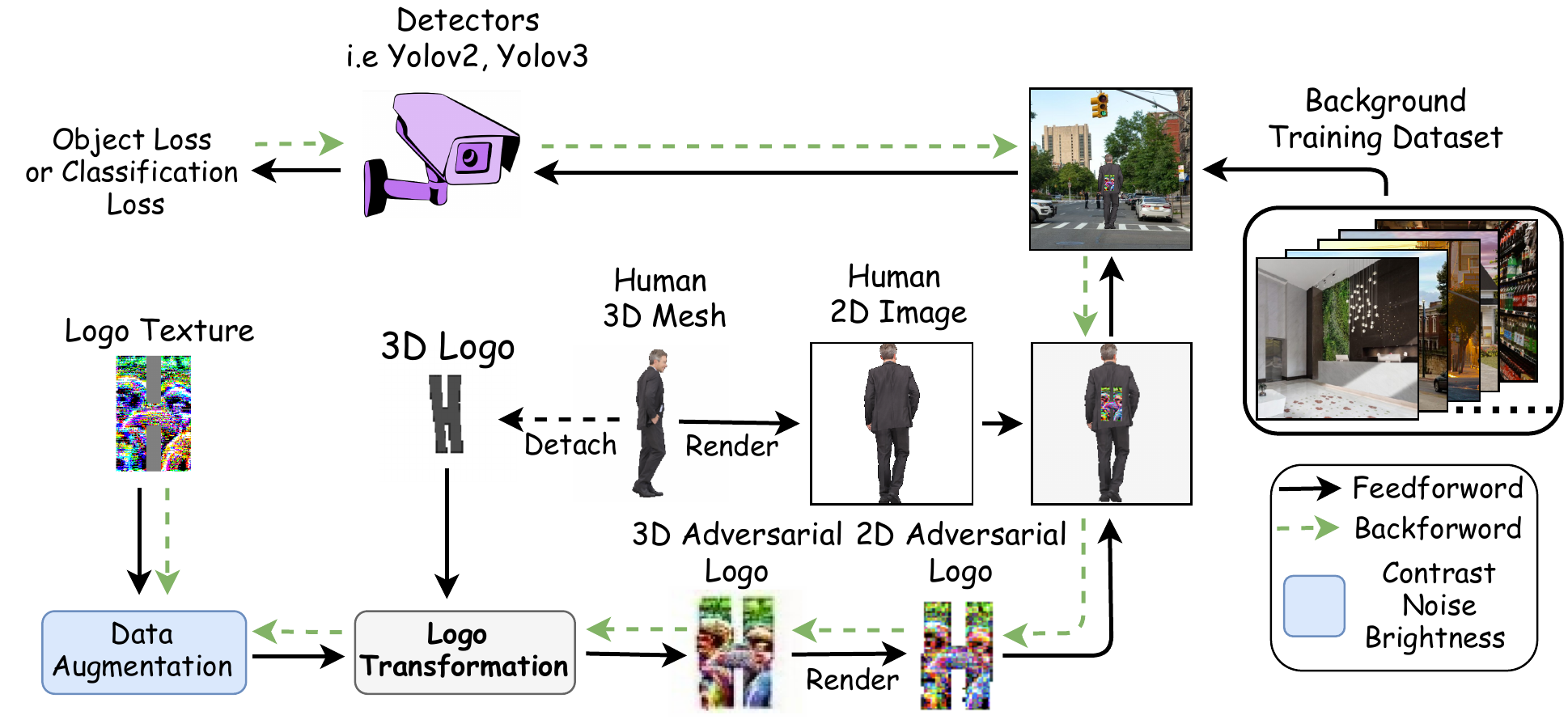}
\caption{The overall framework of our work. We start by choosing a logo texture image, which could be controlled to vary its brightness, contrast, and noise as augmentation. Then we construct textures map that only maps into specific regions over each person's mesh to form 3D adversarial logos. We next apply a differentiable renderer to render the mesh together with its adversarial logo to synthesize the 2D person and 2D adversarial logo. The 2D adversarial logo will then be synthesized with the person image and background images to generate training/testing images. Eventually, the training/testing images are fed into a target object detector for adversarial training/testing.}
\label{fig:Training}
\end{figure*}

In this section, we seek a concrete solution to the 3d adversarial logo attack, with the following goals in mind:

\begin{itemize}

\item [$\bullet$] The adversarial training is \textit{differentiable}: we can modify the source logo via end-to-end loss back-propagation. The major challenge is to replace a traditional discrete render into a differentiable one and to update corresponding texture maps over each mesh.
    
\item [$\bullet$] The 3D adversarial logo is \textit{universal}: for every distinct human mesh, we hope to stitch the 3D adversarial logo generates from the identical 2D logo texture. During both training and testing, we only modify the rendered texture of that logo to have it change concurrently with the human mesh textures. 
\end{itemize}

Our 3D adversarial logo attack pipeline is outlined in Figure \ref{fig:Training}. In the training process, we first define a target logo texture with given shape in the 2D domain ($e.g.$, the character ``H"), and perturb the logo texture by random noise, contrast, and brightness. Then we map the logo texture to 3D surfaces to form 3D adversarial logos on different meshes. Then each human mesh and its 3D adversarial logo are together rendered into a 2D person image associated with its 2D adversarial logo\footnote{We refer to the 2D adversarial logo as the 2D image counterpart that the 3D adversarial logo is rendered into.}. 
These images of person with a logo are further synthesized with background images,
after which we stream these synthesized images into the object detector for adversarial training. 

Due to the end-to-end differentiability, the training process updates the 3D adversarial logo via back-propagation, and further be back-propagated to the logo texture. Within one epoch the above process will be conducted on all training meshes until all background images are trained with, therefore ensuring the logo's universal applicability to different meshes.

\subsection{Logo Transformation}
\label{sec:logo:transfer}
Different from existing 3D attacks, our network aims at a universal patch attack across different input human meshes. This is achieved by editing the 3D adversarial logo over multiple meshes concurrently. Due to discrete polygonal mesh settings, as well as the high possible degrees of distortions and deformations in different 3D meshes, training one universal adversarial logo is highly challenging. We illustrate our logo transformation strategy below, which offers one explicit construction of textures coordinate map for each 3D logo to generate our 3D adversarial logo. Detailed implementations are included in supplementary materials.

Given the logo texture $\mathcal{S}$ define as an RGB image, in order to convert it into a 3D logo $\mathcal{L}$ that can be edited on a single human mesh, our proposed logo transformation comprises of two basic operations: 
\begin{itemize}
\item [$\bullet$] 2D Mapping ($\mathcal{M}_{\mathrm{2D}}$): Project a 3D logo surface onto the 2D domain $[0,1]^2$ to generate texture coordinate mapping.
\item [$\bullet$] 3D Mapping ($\mathcal{M}_{\mathrm{3D}}$): Extract color information from the logo texture and map color information onto each face over a 3D logo to composite a 3D adversarial logo.
\end{itemize}

The overall logo transformation can be denoted as:
\begin{equation}
  \tilde{\mathcal{L}}= \mathcal{T}_{\mathrm{logo}}(\mathcal{S},\mathcal{L}) = \mathcal{M}_{\mathrm{3D}}(\mathcal{S}, \mathcal{M}_{\mathrm{2D}}(\mathcal{L})) 
  \label{logo transform: overall}
\end{equation}

With the logo transformation, the chosen 2D logo shape is mapped to 3D logo $\mathcal{L}$ on each distinct human mesh to form the 3D adversarial logo $\tilde{\mathcal{L}}$. By leveraging a differentiable renderer (to be discussed in Section \ref{sec:differentiable:renderer}), when rendering the 3D adversarial logos into adversarial logo images, the updates will be back-propagated from those images to 3D adversarial logos and thereby all the update information are aggregated to the logo texture. In our work, the logo transformation is only constructed once and we fix the texture coordinate map for each human mesh. When forwarding to the detector, the color of our 3D adversarial logos is obtained via distinctive texture coordinate maps and hence we can update all 3D adversarial logos based on logo texture in synchronization. 

\begin{figure}[t]
\centering
\includegraphics[width=1\linewidth]{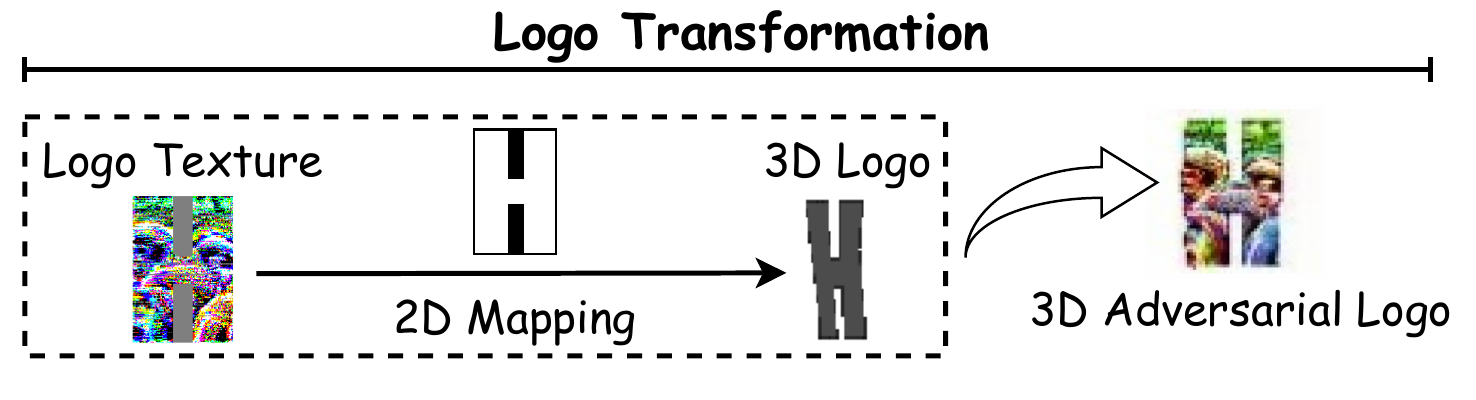}
% \vspace{-2mm}
\caption{Our Logo Transformation scheme. 3D logos are detached from the person mesh as submeshes, then they are mapped to the logo texture via texture mapping(2D Mapping) we construct. 3D adversarial logos are generated by assigning color information onto 3D logos from the logo texture.}
\label{fig:logotransfer}
\vspace{-5mm}
\end{figure}

\subsection{Differentiable Renderer}
\label{sec:differentiable:renderer}
A differentiable renderer can take meshes as input and update the mesh's texture via back-propagation for different purposes. Our work is built upon a specific renderer called Neural 3D Mesh Renderer \cite{kato2018renderer}, while any other differentiable renderer shall serve our goal here.

The Neural Renderer proposed in \cite{kato2018renderer} generates $q\times q \times q$ color cubes for each face over the mesh. By adopting an approximate rasterized gradient, where piece-wise constant functions are approximated via linear interpolations, as well as centroid color sampling, the renderer is capable of editing mesh's texture through back-propagation. We apply the Neural Renderer to render the 3D logo output from the logo transformation, into various 2D adversarial logos, and meanwhile, render the 3D human mesh. The last step is to attach an (augmented) 2D adversarial logo to the corresponding rendered 2D person image, and then synthesize with a real background image, yielding the final 2D adversarial image that aims to fool the object detector. During back-propagation, the update in 2D adversarial logo images will be fed back to the 3D adversarial logo, and eventually back to the initial input of logo texture, thanks to the renderer's differentiability.

\subsection{Training Against 3D Adversarial Logo Attacks}
\label{sec:loss}
The aim of our work is to generate a 3D adversarial logo on human mesh and the human mesh with this 3D adversarial logo can fool the object detector when it is rendered into a 2D image. We next discuss how we compose the training loss to achieve this goal.
\paragraph{Disappearance Loss}
To fool an object detector is to diminish the confidence within bounding boxes that contain the target object. Such that it cannot be detected. We exploit the disappearance loss \cite{eykholt2018physical}, which takes the maximum confidence of all bounding boxes that contain the target object as the loss:
\begin{equation}
    \mathrm{DIS}(\mathcal{I}, y) = \mathrm{Conf}_{\mathrm{max}}(\mathcal{O}(\mathcal{I}, y)),
    \label{eq:loss:DIS}
\end{equation}
where $\mathcal{I}$ is the image streamed into the object detector and $y$ is the object class label, $\mathcal{O}$ is the object detector that outputs bounding box predictions, and $\mathrm{Conf}_{\mathrm{max}}$ calculate the maximum confidence among all the bounding boxes.

\paragraph{Total Variance Loss}
To further smooth the predictions over augmented 2D adversarial logos and avoid inconsistent predictions, a total variance loss is enforced \cite{sharif2016accessorize}.
\begin{equation}
  \small
  \begin{split}
  \mathrm{TV}(\tilde{\mathcal{L}}) = \sum_{i,j} (|\mathcal{R}(\tilde{\mathcal{L}})_{i,j}-\mathcal{R}(\tilde{\mathcal{L}})_{i,j+1}|  + |\mathcal{R}(\tilde{\mathcal{L}})_{i+1,j}-\mathcal{R}(\tilde{\mathcal{L}})_{i,j}|)  
  \end{split}
  \label{eq:loss:TV}
\end{equation}
where $\mathcal{R}$ is denoted as the differentiable renderer in Section \ref{sec:differentiable:renderer}. We apply such a notation to emphasize the loss is computed over pixel values of 2D adversarial logos, and $\mathcal{R}(\tilde{\mathcal{L}})_{i,j}$ is one specific pixel value at coordinate $(i,j)$. This loss is added to improve physical realizability.

% \vspace{0.3em}
The overall training loss we are minimizing is composed of the above two losses ($\lambda_{\mathrm{DIS}}$ and $\lambda_{\mathrm{TV}}$ are the hyper-parameters):
\begin{equation}
  \mathcal{L}_{\mathrm{adv}} = \lambda_{\mathrm{DIS}} \mathrm{DIS}(\mathcal{I}, y) + \lambda_{\mathrm{TV}}\mathrm{TV}(\tilde{\mathcal{L}}) 
  \label{eq:loss:overall}
\end{equation}

\section{Experiments and Results}

\begin{figure*}[t]
\centering
\includegraphics[width=1\linewidth]{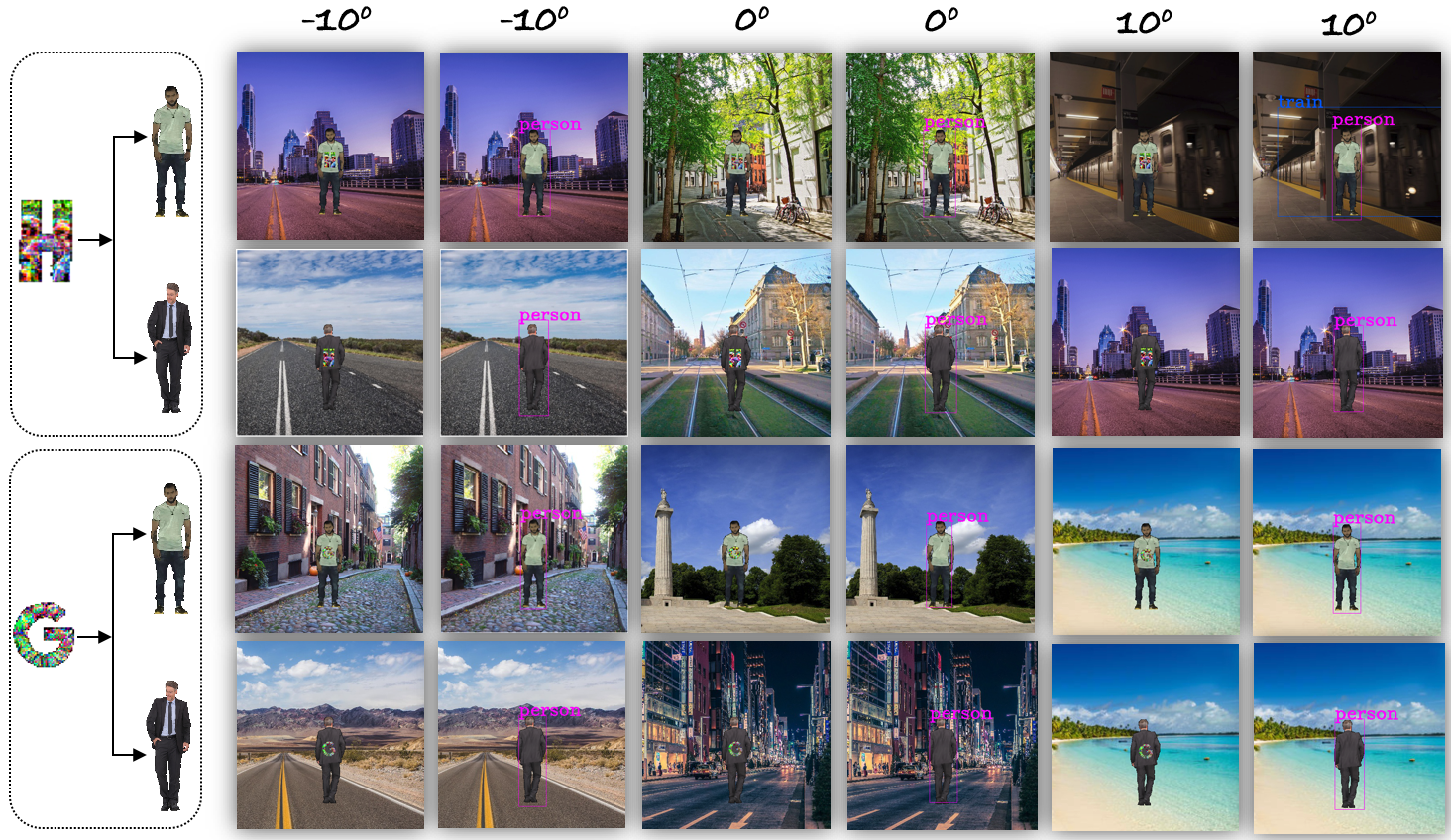}
\caption{Examples of our adversarial attack to YOLOV2. We perform multi-angle training over three meshes. To justify our work can attack in different rendering angles, we attach our adversarial logo into different positions on different human meshes (one is in the front while the other is at the back).  We display the results of logo ``G" (first row) and logo ``H" (second row) under three different views (-10, 0, 10 degrees). 
The result in three different angle views conceptually verifies that our adversarial logos can prevent the objector detector from recognizing the person in different poses, even if perturbed by 3D rotations.}
\label{fig:res}
% \vspace{-2mm}
\end{figure*}

\subsection{Dataset Preparation}

\begin{figure}[h]
\centering
\includegraphics[width=1\linewidth]{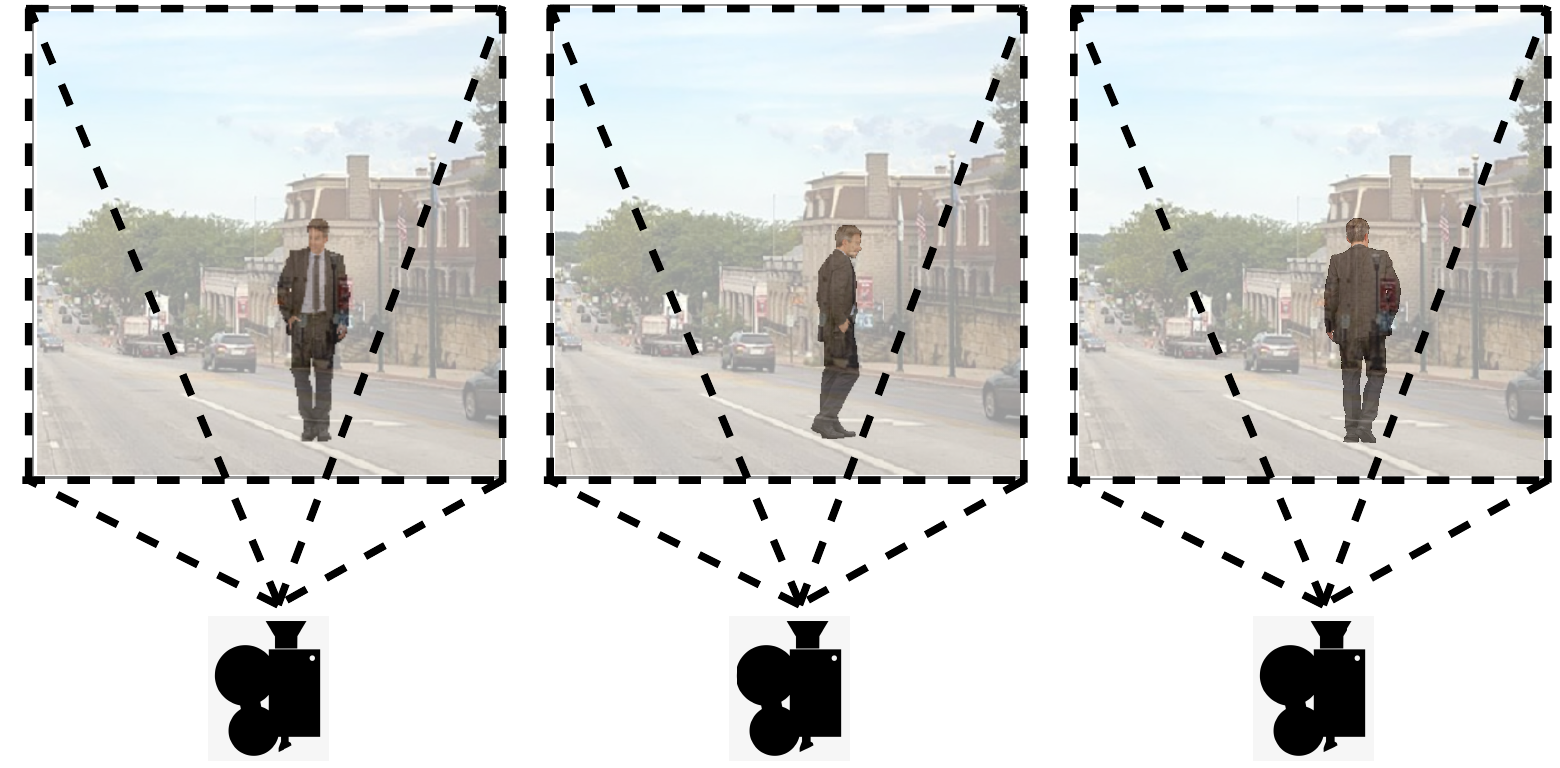}
\caption{An angle view setting example. From left to right: $-90$ degree, $0$ degree and $+90$ degree for one background image and one human model.}
\label{fig:view}
\end{figure}
\paragraph{Generating Multiple Angle views}

We generate 2D adversarial images via the Neural Renderer under varying angle views. Specifically, we pick up one specific angle view as our benchmark view of $0$ degree. We fix the rendering camera view and rotate the 3D human model. We denote ``$+$'' for counterclockwise rotation while ``$-$'' for clockwise rotation. Figure $\ref{fig:view}$ presents an example of our angle view settings. When synthesized with background images, we always assume the fixed camera view captures the background, so there is no further cropping, translation nor rotation operation for background images.

% We generate 2D adversarial images via the Neural Renderer under varying camera views. Specifically, we pick up one specific camera view as our benchmark view of $0$ degree. We assume that our rendering camera can be rotated along the horizontal equator of a sphere starting from the benchmark view we predefined as above. The synthesized 2D background images with humans and adversarial logos rendered from those camera views, where camera positions are sampled along the circumference of the equator. The top row of Figure \ref{fig:intro} presents the rendering results under different camera view settings (-10, 0 and 10 degrees). {\color{red} When synthesized with background images, we  always assume the background is fixed under every camera view, and only human mesh models are rotating under different views. (Can we add a figure here to explain this?)}

\paragraph{Background Images}

For training backgrounds, we crawl a set of real photos from Google search, using keywords \textit{street, avenue, park}, and \textit{lawn}. We manually inspect all photos and discard those that contain a person or other object distractors. We end up with 312 ``clean background''-style images that are used for training. There are multiple synthesized images generated per background image, depending on how many angle-views we sample. For most of our experiments (except for single-angle training, see Section \ref{sec:single:multiple}), we sample over 21 views, leading to a training set of size 6,741. We scale up testing background images by sampling over images from MIT places dataset \cite{zhou2017places} with the same criterion, generating 32,000 ``clean background``-style images for testing.

\paragraph{Mesh Model Data} For the mesh data, we select three publicly available 3D human models\footnote{\url{https://www.turbosquid.com/3d-models/water-park-slides-3d-max/1093267} and \url{https://renderpeople.com/free-3d-people/} contain the source mesh.} with complete textures and coordinate maps. We edit each mesh to select faces that form our 3D logos under given 2D shape contours. Afterward, we process every 3D logo via \textit{OpenMesh 8.0}\,\footnote{\url{https://www.openmesh.org/}} to extract its centroid coordinates for our logo transformation step. 
 
\subsection{Implementation Details}
\label{sec:experiments:settings}
All experiments are implemented in PyTorch 1.0.0, along with the Neural Renderer PyTorch implementation\footnote{\url{https://github.com/daniilidis-group/neural_renderer}}. We choose the default hyperparameters in the Neural Renderer as: elevation is 0, camera distance is 2.0, ambient light is 1.0, light direction is 0.0 and the cubic size $q$ in Section \ref{sec:differentiable:renderer} is 4. For data augmentation, we add the contrast uniformly generated from 0 to 1, the brightness uniformly generated from 0 to 1, and the noise uniformly generated from $-0.1$ to $0.1$. All three are added pixelwise. The training is conducted on one Nvidia GTX\,1080TI GPU. The default optimizer is Adam, and the learning rate is initialized as 0.03 and decays by a factor of 0.1 every 50 epochs. 

During all experiments, $\lambda_{\mathrm{DIS}}=1.0$ and $\lambda_{\mathrm{TV}}=2.5$ in \eqref{eq:loss:overall} respectively. The batch size of background images is set as 8, while the batch size of synthesizing rendering meshes is set to 1. The total training epochs in the single-angle training and multiple-angle training (explained in Section \ref{sec:single:multiple}) are 100 and 20 respectively.

The default object detectors used are YOLOv2 \cite{redmon2017yolo9000} and YOLOv3 \cite{redmon2018yolov3}, with confidence thresholds set to 0.6 and 0.7 respectively. Shapes of logo texture we mostly exhibit are characters of ``G" and ``H", while more character shapes are investigated in our study as well. 
 
\subsection{Experiment Results and Analysis}
\subsubsection{Single- and Multi-angle Evaluation}
\label{sec:single:multiple}
\paragraph{Single-angle training} We first apply our 3D adversarial attack pipeline over single-angle rendering images as \textit{single-angle training}. We illustrate the main idea using $0$ degree, for example. Given a mesh, we synthesize our 2D images with background images in the training set under the $0$ degree angle view. Then we train over those images to obtain our adversarial logos and logo texture. The testing images are rendered under the same $0$ degree view and synthesized with test backgrounds. The attack success rate denotes the ratio of testing samples that that target detector misses the person. Table \ref{table:angle}'s first column compares the results: as a proof-of-concept, our proposed attack successfully compromises both YoloV2 and YoloV3 (appear to be relatively more robust), under both ``H" and ``G" logo shapes.

\paragraph{Multi-angle training} We then extend to a joint 3D view training setting called \textit{multi-angle training}. We uniformly sample the degrees between [-10, 10] with one-degree increment, leading to 21 discrete rendering views. Under this setting, both the training set and test set are enlarged by a multiplier of $21$. We compute our multi-angle success rate by considering averaging the success rates across all $21$ views. Results are summarized in the second column of Table \ref{table:angle}. As can be seen in table \ref{table:angle}, a lower success rate implies that the multi-angle attack is more challenging compared to the single-angle attack, where no perturbation is considered.

The numbers we report in Table \ref{table:angle} are consistent with our visual results. Some selective images of our multi-angle training are shown in Figure \ref{fig:res}. 
As one could observe, our adversarial logo can mislead the pre-trained object detector and make the person unrecognizable under both logo ``G" and logo ``H".

% \begin{table}[ht]
% \begin{center}
% \caption{Results of logos ``G" and ``H" in single-angle and multiple-angle training. The baselines are by applying detectors to the synthesized human images without adversarial logos.}
% \label{table:angle}
% \begin{threeparttable}
% \resizebox{0.48\textwidth}{!}{
% \begin{tabular}{c|c|c|c|c|c|c}
% \hline
% \multirow{2}{*}{Object Detector} & \multicolumn{6}{c}{\textbf{Attack Success Rate}} \\ \cline{2-7} 
%  & YoloV2 (Baseline) & YoloV3 (Baseline) & YoloV$2_H$ & YoloV$3_H$ & YoloV$2_G$ & YoloV$3_G$ \\
%  \hline
% single angle view & 0.01 & 0.01 & 0.86 & 0.91 & 0.79 & 0.60 \\
%  \hline
% multiple angle views & 0.01 & 0.01 & 0.88 & 0.74 & 0.67 & 0.41 \\
% \hline \hline
% \end{tabular}}
% \end{threeparttable}
% \end{center}
% \end{table}

\begin{table}[ht]
\begin{center}
\caption{Results of logos ``G" and ``H" in single-angle and multiple-angle training. The baselines are by applying detectors to the synthesized human images without adversarial logos.}
\label{table:angle}
\begin{threeparttable}
\resizebox{0.48\textwidth}{!}{
\begin{tabular}{c|c|c}
\hline
\multirow{2}{*}{Object Detector} & \multicolumn{2}{c}{\textbf{Attack Success Rate}} \\ \cline{2-3} 
& single angle view & multiple angle views \\
\hline
YoloV2 (Baseline) &  0.01 & 0.01  \\
YoloV3 (Baseline) & 0.01 & 0.01 \\
YoloV$2_H$ & 0.86 & 0.88 \\
YoloV$3_H$ & 0.91 & 0.74 \\
YoloV$2_G$ & 0.79 & 0.67 \\
YoloV$3_G$ & 0.60 & 0.41 \\
\hline \hline
\end{tabular}}
\end{threeparttable}
\end{center}
\end{table}

\subsubsection{Attack to unseen angle views}

\begin{figure}[t] 
\centering
\includegraphics[width=1\linewidth]{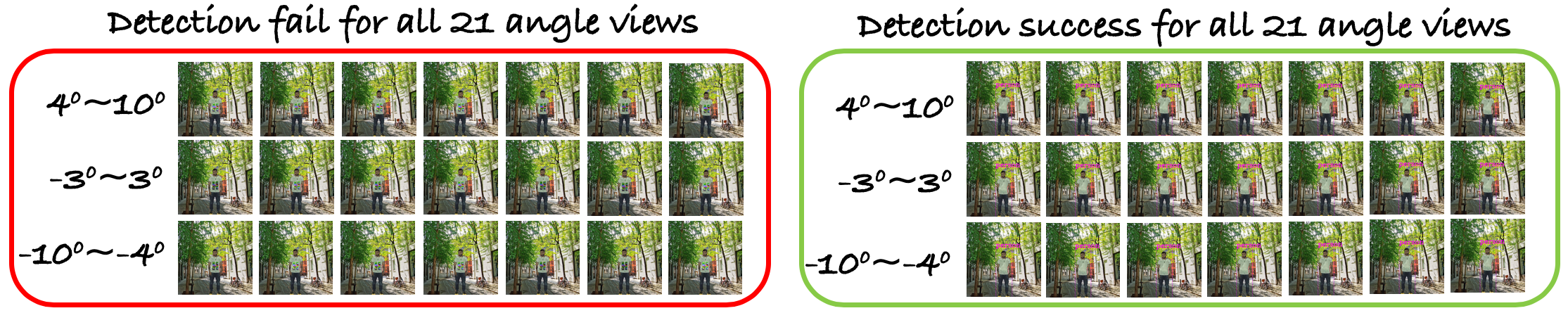}
\caption{A success attack under unseen views with single-angle training. Our 3D adversarial logo can fool YOLOV2 detector under all $21$ angle views, $20$ of which are never seen by the detector when training.}
\label{fig:oneforall}
\end{figure}

\paragraph{Single-angle training with Multi-angle attack}
To prove our method is robust against 3D rotations, we conduct a multi-angle attack with single-angle training. We first train at $0$ degree, but use $21$ views ($[-10,10]$ degree) to attack the detector. Results shown in Figure \ref{fig:2d3d}(b) proves our method is stable against small model angle perturbations. Figure \ref{fig:oneforall} provides an example where our 3D adversarial logo hides the person from the detector. Though indistinguishable from human eyes, tiny perturbations in images cannot be underestimated in adversarial attacks \cite{zhang2019making,azulay2018deep,engstrom2019exploring}. Nevertheless, our method is not affected by pixel-level changes that could collapse other 2D patch adversarial works.

% \vspace{-3mm}
\paragraph{Multi-angle training with unseen-angle attack} 

\begin{figure}[ht]
\centering
\includegraphics[width=1\linewidth]{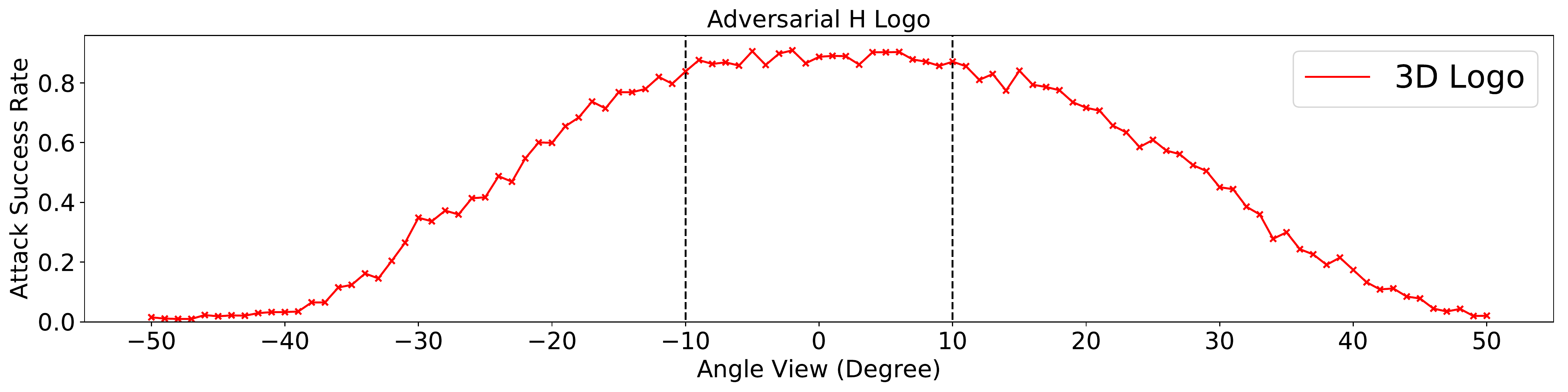}
\caption{The attack success rate for each angle view with multi-angle training. The detector is YOLOV2 and the training views ($[-10,10]$ degree) are highlighted within dashed lines. There exists a massive performance drop when angle view change is relatively large.}
\label{fig:multi-wider}
% \vspace{-10mm}
\end{figure}

\begin{figure}[ht]
\centering
\includegraphics[width=1\linewidth]{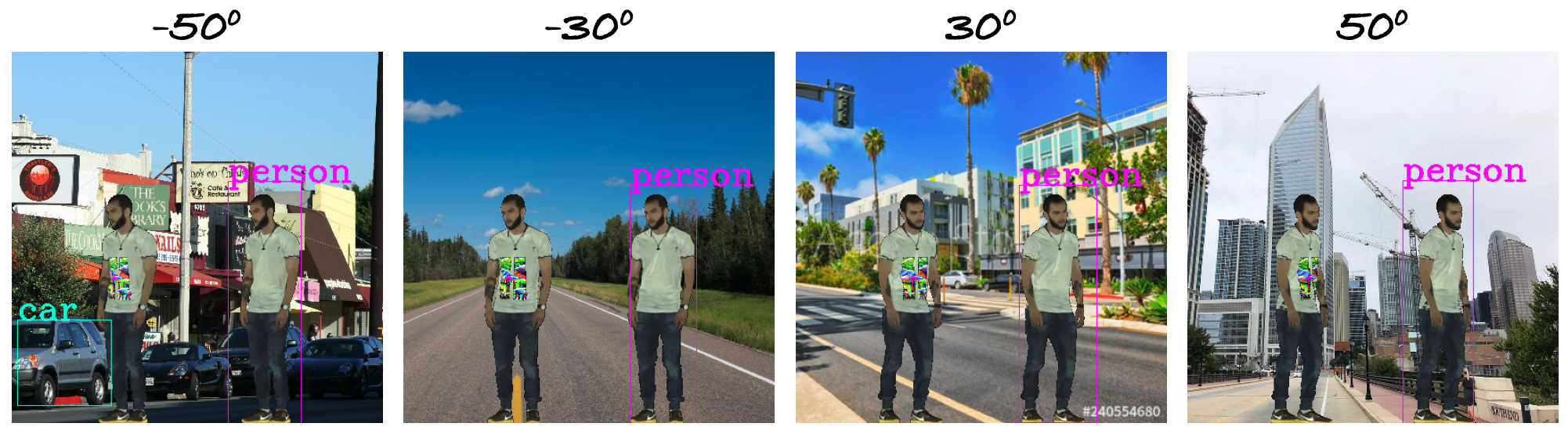}
\caption{Our attack result for YOLOV2 under unseen angle views. When the camera rotates drastically, part of our adversarial logo disappeared from the rendered image, and our attack fails when the unseen view is 50 degree.}
\label{fig:unseen3050}
% \vspace{-4mm}
\end{figure}

We extend our experiments to test robustness under more angle views. After multi-angle training with $21$ views in $[-10,10]$ degree, we attack the detector using all $101$ angle views in $[-50,50]$ degree. Figure  \ref{fig:multi-wider} is plotted based on our success attack rate overall test images and we provide examples of unseen view attacks in Figure \ref{fig:unseen3050}. The plot in Figure \ref{fig:multi-wider} reveals the limitations of our works. When our rendering view is away from training views, we observe a decaying curve that converges to $0$ eventually. One plausible explanation is that clipping of our adversarial logos in rendering pipeline leads to loss of information from our logo textures, thereby severely affects our performance.

\subsubsection{Comparison with adversarial patch attacks}

For comparison fairness with previous adversarial patch attacks (cf. \cite{thys2019fooling,chen2018shapeshifter,eykholt2018physical,wu2019making}), we perform single-angle training in our pipeline and conduct conventional 2D patch adversarial attack \cite{thys2019fooling} as follows: i) We apply masks to generate 2D patches that obtain identical shape as our logo texture. ii) Different from our 3D adversarial attack, We add 2D perturbations (translation and rotation) to optimize the performance of 2D patch adversarial attacks. iii) The comparison is to perform the same testing setup, but replace the logo trained in our proposed method with the masked patch trained via state-of-the-art. In other words, the major difference is whether 3D perturbations are considered during training.

\begin{figure}[ht]
\centering
\includegraphics[width=1\linewidth]{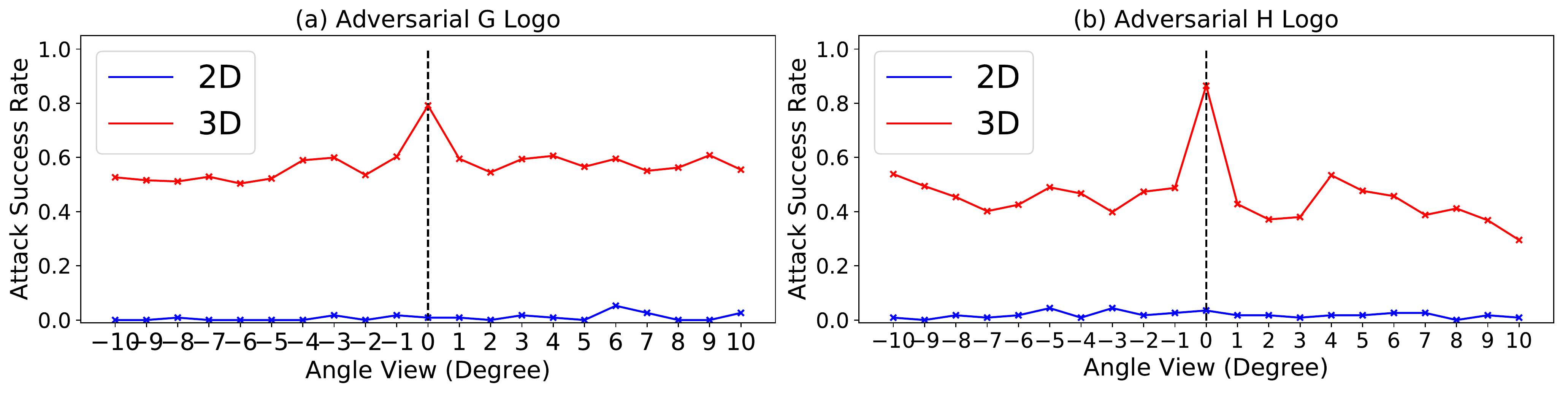}
\caption{The attack success rate for each angle view in $[-10,10]$ degree. We test performance between the 2D adversarial patch attack with our 3D adversarial logo attack on YOLOV2. The dashed line emphasizes that we only train in a single-angle but test in multiple angle views.}
\label{fig:2d3d}
% \vspace{-8mm}
\end{figure}

We compare the performance of two schemes under the multi-angle attack. We synthesize our test images with $38$ test background images under $21$ different views ($[-10,10]$ degree), by applying the identical rendering scheme for both 2D and 3D methods (by replacing logo texture with 2D adversarial patch when performing 2D patch multi-angle testing). Only by this procedure can we ensure the rendered logos share the same positions and same perturbation under model angle changes. We compare attack success rate (defined in Section \ref{sec:single:multiple}) between two methods and report the result in Figure $\ref{fig:2d3d}$. We observe that our method outperforms the conventional 2D patch adversarial attack scheme. Even at the training view (0 degree), the 2D method fails, due to the distortion when our logo is dropped on clothes via rendering. This experiment indicates our adversarial logos can naturally shift as the physical scene changes. We do not require the patch is either center-aligned or at a fixed position, whereas the method in \cite{thys2019fooling} obtains a position-sensitive adversarial patch.

\subsubsection{Shape adaptivity}

\begin{figure}[ht]
\centering
\includegraphics[width=0.8\linewidth]{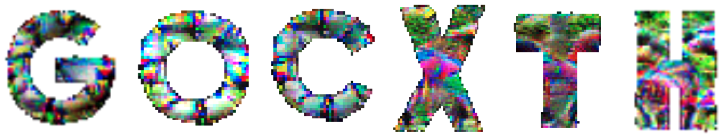}
\caption{Our shape gallery. Trained logo textures from left to right are: ``G",``O",``C",``X",``T",``H".}
\label{fig:logo}
% \vspace{-8mm}
\end{figure}

\begin{figure}[ht]
\centering
\includegraphics[width=1\linewidth]{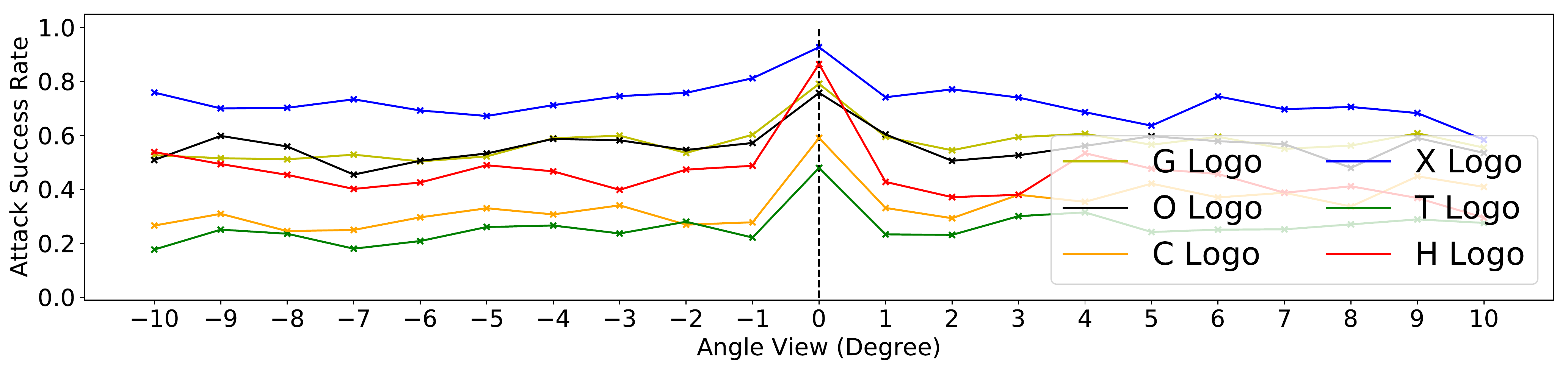}
\caption{The attack success rate among six different logo shapes. We conduct a single-angle training (0 degree) and test under $21$ different views, with only the shape of logo texture altered.}
\label{fig:logo:plot}
% \vspace{-8mm}
\end{figure}

Although our 3D adversarial logo attack is not restricted to the shape of the logo, the result in different shapes reveals that numerical differences originated from different shapes are not negligible, as seen in Table \ref{table:angle} and Figure \ref{fig:logo}. We select six different shapes (character  ``G",``O",``C",``X",``T",``H") as contour of our logo texture. The motivation is that the first three characters contain curves and complicated 2D contours while the latter three consist of parallel and regular contours. We believe these characters can cover all common cases when designing a new shape of logo texture. Figure \ref{fig:logo:plot} compare their performance when performing multi-angle attack under single-angle training. It can be seen that the more regular and symmetry the shape is, the better attack success rate is attained. However, this does not hold for the logo of the character ``T". A possible explanation is that the symmetry along a horizontal axis might be most crucial to deceive the detector. Both ``C" and ``G" outperform ``T", making the letter ``T" the worst result among all  logos.

% \vspace{-2mm}

\subsubsection{Attack to unseen human mesh}
\begin{figure}[ht]
\centering
\includegraphics[width=1\linewidth]{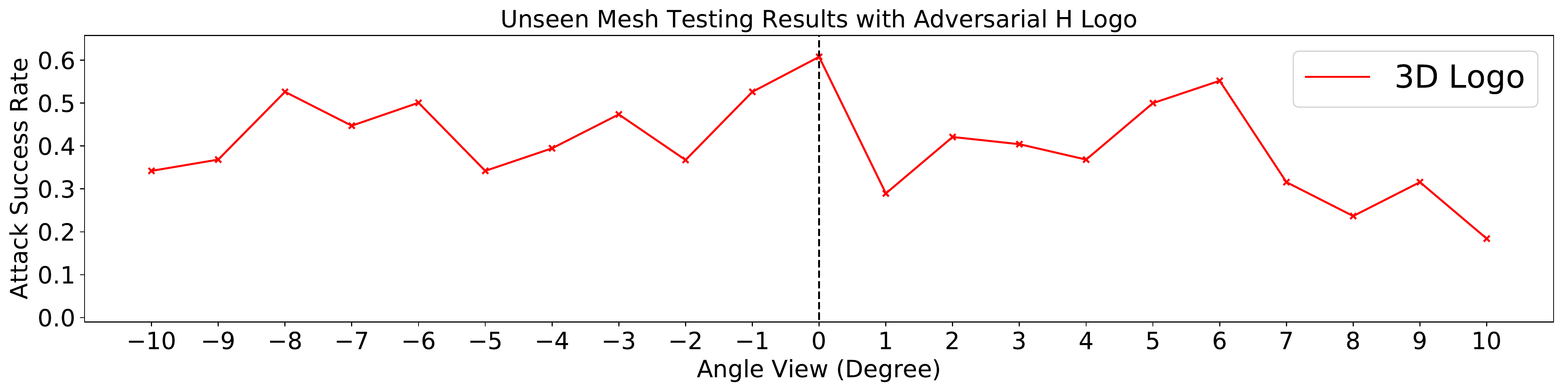}
\caption{Attack success rate under the unseen mesh. We perform single view training (0 degree) and test under 21 different views. Compare to Figure \ref{fig:2d3d}, we test our result over a new mesh which is not seen during training process.}
\label{fig:cross:mesh}
\vspace{-4mm}
\end{figure}

One goal of our framework is to transfer one logo texture across all 3D logos on distinctive meshes, and only the logo texture will be updated via back-propagation. This setting brings us joint mesh training and promising one-for-all generalization to attack the detector in different meshes. In Figure \ref{fig:cross:mesh}, we justify our joint-training maintains consistent performance over unseen meshes. We use two meshes for training and use the third mesh(The woman) to test our generated adversarial logo performance. The mean attack success rate is $40$\%. Although we did not observe the same attack success rate compared to same-mesh training, our method remains its potential to attack in more generalizable cases where logos are shown on different humans.

\subsubsection{Attack to unseen detector}
\begin{figure}[ht]
\centering
\includegraphics[width=1\linewidth]{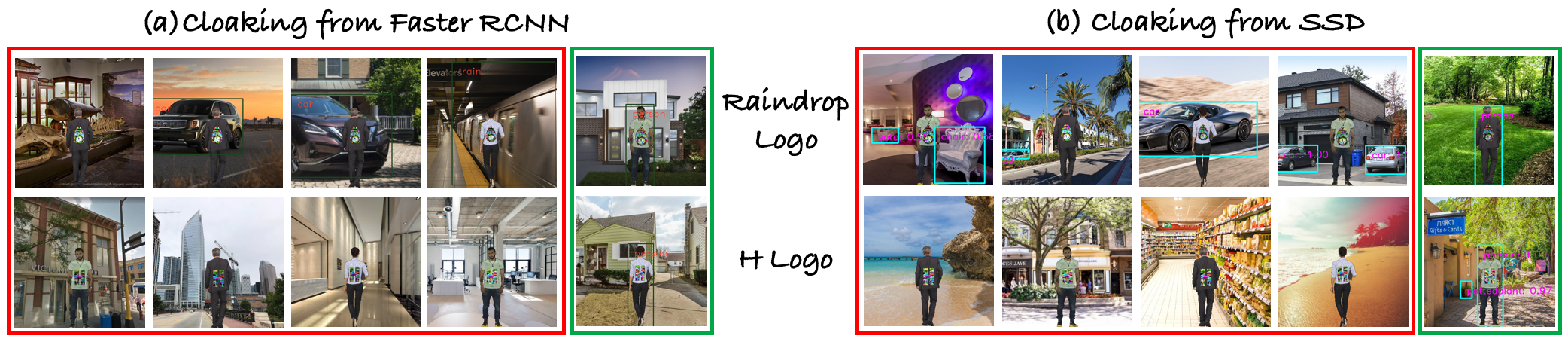}
\caption{Examples of our 3D adversarial logos in unseen detectors. Left: attack in Faster-RCNN; Right: attack in SSD. The first eight images in \textcolor{red}{red} box is success cases while two right-end images in \textcolor{green}{green} box are failure cases.}
\label{fig:unseendetector}
\vspace{-3mm}
\end{figure}

To test our generalizability across detectors, we choose Faster-RCNN (Faster Region-based Convolutional Neural Networks in \cite{shaoqing2015fasterRCNN}) and SSD(Single Shot Detector in \cite{wei2015SSD}) as our targets. We train our adversarial logos on YOLOV2 under $[-10,10]$ degree of views and feed test images into two unseen detectors under $[-50,50]$ degree views. We achieve 42\% average attack success rate over SSD while the number is 46\% over Faster-RCNN. Successful and failure examples for two detectors are shown in Figure \ref{fig:unseendetector}, respectively. Note that in \cite{thys2019fooling}, the transferability of 2D adversarial patch attacks to unseen detectors is often in jeopardy. In comparison, our 3D adversarial logo appears to transfer better across unseen detectors, despite not being specifically optimized.

\section{Conclusion}

We have presented our novel 3D adversarial logo attack on human meshes. We start from the logo texture with designated shape and diverge to 2D adversarial logos that are naturally rendered from 3D adversarial logos on human meshes. Due to differentiable renderer, the update back to the logo texture image is shape-free, mesh-free, and angle-free, leading to a stable attack success rate under different angle views with different human models and logo shapes. Our method enables a fashion-designed potential in the realistic adversarial attack. In the future, we hope to justify the feasibility of our method in the physical-world by examining the printability of our adversarial logos.

{\small
\bibliographystyle{ieee_fullname}
\bibliography{egbib}
}

\end{document}

%%%%%%%%% TITLE
\title{Can 3D Adversarial Logos Cloak Humans?}

\author{First Author\\
Institution1\\
Institution1 address\\
{\tt\small firstauthor@i1.org}
Second Author\\
Institution2\\
First line of institution2 address\\
{\tt\small secondauthor@i2.org}
}

\maketitle

\section{Explicit Mappings of logo transformation}

We now give the explicit formula of both 2D mapping and 3D mapping we defined back in Section 3.1. Recall we construct texture-to-logo mapping via:
\begin{equation}
  \tilde{\mathcal{L}} = \mathcal{T}_{\mathrm{logo}}(\mathcal{S}, \mathcal{L}) = \mathcal{M}_{\mathrm{3D}}(\mathcal{S}, \mathcal{M}_{\mathrm{2D}}(\mathcal{L})) 
  \label{eq:supp:logo:transofrmation}
\end{equation}
where $\mathcal{L}$ is the input 3D logo and $\mathcal{S}$ is the logo texture. $\mathcal{S}$ is defined on a 2D image of size $h\times w$, which only contains color information within the contour of the given shape.

\paragraph{2D Mapping}
In the 2D mapping, we first extract the centroid coordinates $x, y, z$ of each face $f\in \mathcal{L}$ and then we scale it into the bounding box of the 3D logo, yielding the relative position coordinate $(\hat{x}, \hat{y}, \hat{z})$ in the set $\mathcal{R}:=[0,1]^{3}$. The mapping $\mathcal{M}_{\mathrm{2D}}$ thus can be written as:
\[  \mathcal{M}_{\mathrm{2D}}(f)= (\hat{x},\hat{y}), \quad \forall f\in \mathcal{L}, \]
which refers to the texture coordinate map from the 3D logo mesh to the relative 2D space $D:=[0,1]^2$. Given the shape of $\mathcal{S}$, if $\hat{x}, \hat{y}$ locate outside the boundary of $\mathcal{S}$ in the 2D space, it will be constrained inside the boundary via finding its nearest neighbor within. 

\paragraph{3D Mapping}

Given the 2D Mapping $\mathcal{M}_{\mathrm{2D}}$, we can retrieve color information of each face $f\in \mathcal{L}$ from the texture logo by scaling from $\mathcal{D}$ to $[0,w]\times[0,h]$. The 3D Mapping binds a texture cube of each face $f$ on 3D logo $\mathcal{L}$ via 2D Mapping to form the face $\tilde{f}$ over our 3D adversarial logo $\tilde{\mathcal{L}}$. Namely,
\[ \tilde{f} = \mathcal{M}_{\mathrm{3D}}(\mathcal{S}, \mathcal{M}_{\mathrm{2D}}(\mathcal{L})) = (\mathcal{S}(w\hat{x},h\hat{y})\cdot\mathbf{1},f),\quad \forall f\in \mathcal{L}.  \]

Here $\mathcal{S}(w\hat{x},h\hat{y})\in\mathbb{R}^3$ refers the color of the corresponding pixel in the logo texture $\mathcal{S}$, and $\mathbf{1}$ is a $q\times q\times q$ tensor with each element $1$.
Then each texture cube is sent to the differentiable renderer. When back-propagating, the gradient will be progressed back to $\mathcal{S}$ eventually. 

\section{Additional Results}

\subsection{Logo in different sizes}
\begin{figure}[t]
\centering
\includegraphics[width=1\linewidth]{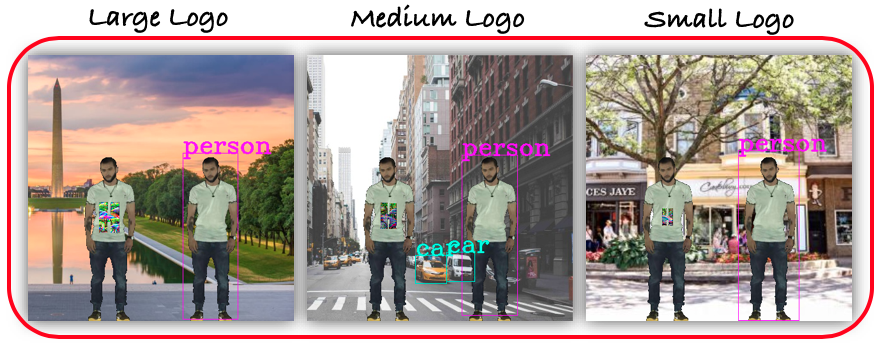}
\caption{Examples of successful attacks for different logo scales. From left to right : large logo (original size) ; medium logo ($2/3$ of original size) ; small logo ($1/3$ of original size).}
\label{fig:logosize}
\end{figure}

\begin{figure}[t]
\centering
\includegraphics[width=1\linewidth]{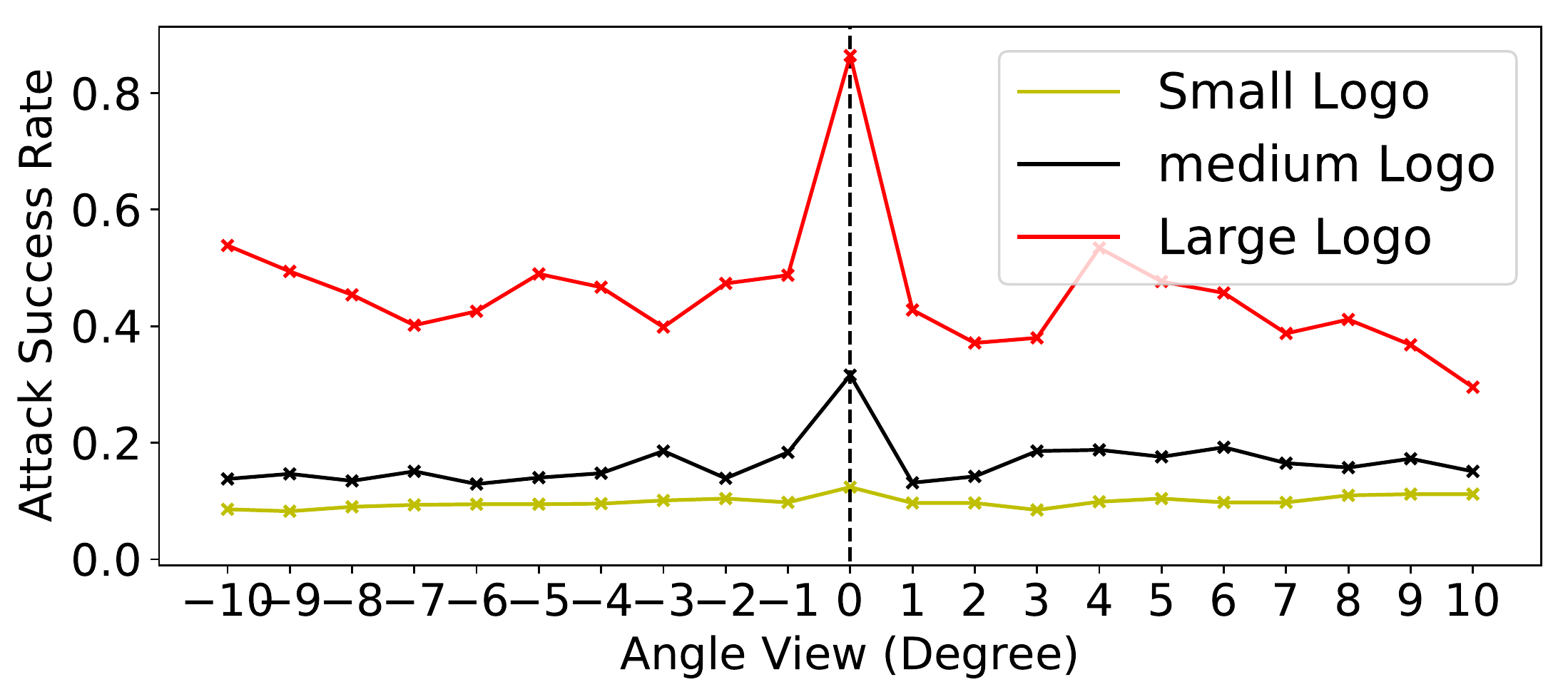}
\caption{Attack performance with regard to different scales of logos. The 3D adversarial logo is trained in single-view (marked in dashed line) and test under multi-view attacks.}
\label{fig:size}
\end{figure}

We perform the test of our adversarial logos in different sizes. Figure \ref{fig:size} reflects some attack results. We perform the same single-view training as did in Section 4.3 and attack it under multiple rendering views ($[-10,10]$ degree). The attack success rates at each individual rendering view are reported in Figure \ref{fig:logosize}. Due to the limited attack region, the attack success rate drops drastically when we shrink the logo.

\subsection{More Irregular Logos}
\begin{figure}[t]
\centering
\includegraphics[width=1\linewidth]{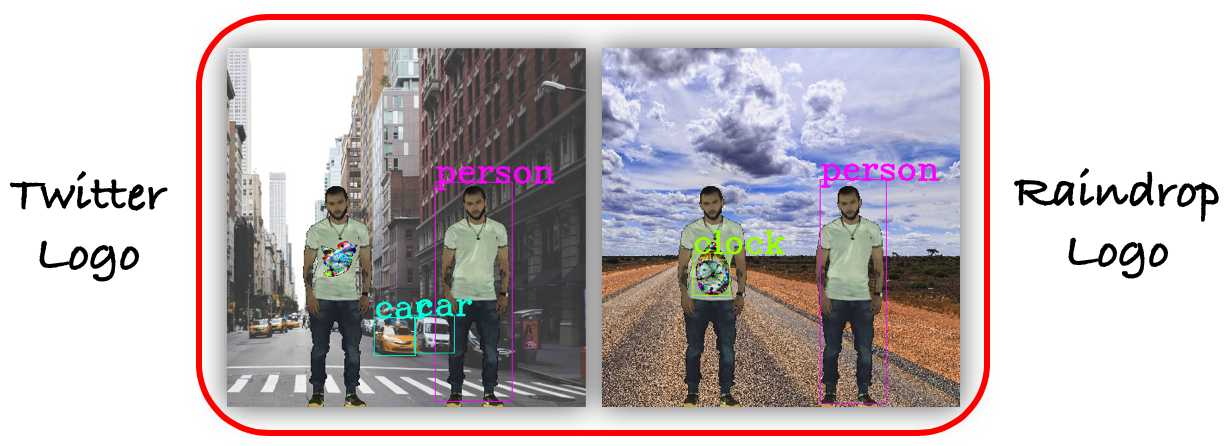}
\caption{Selective deceptions in different logo shapes. Left: Twitter-like logo; Right: Raindrop-like logo.  }
\label{fig:extralogo}
\end{figure}

\begin{figure}[t]
\centering
\includegraphics[width=1\linewidth]{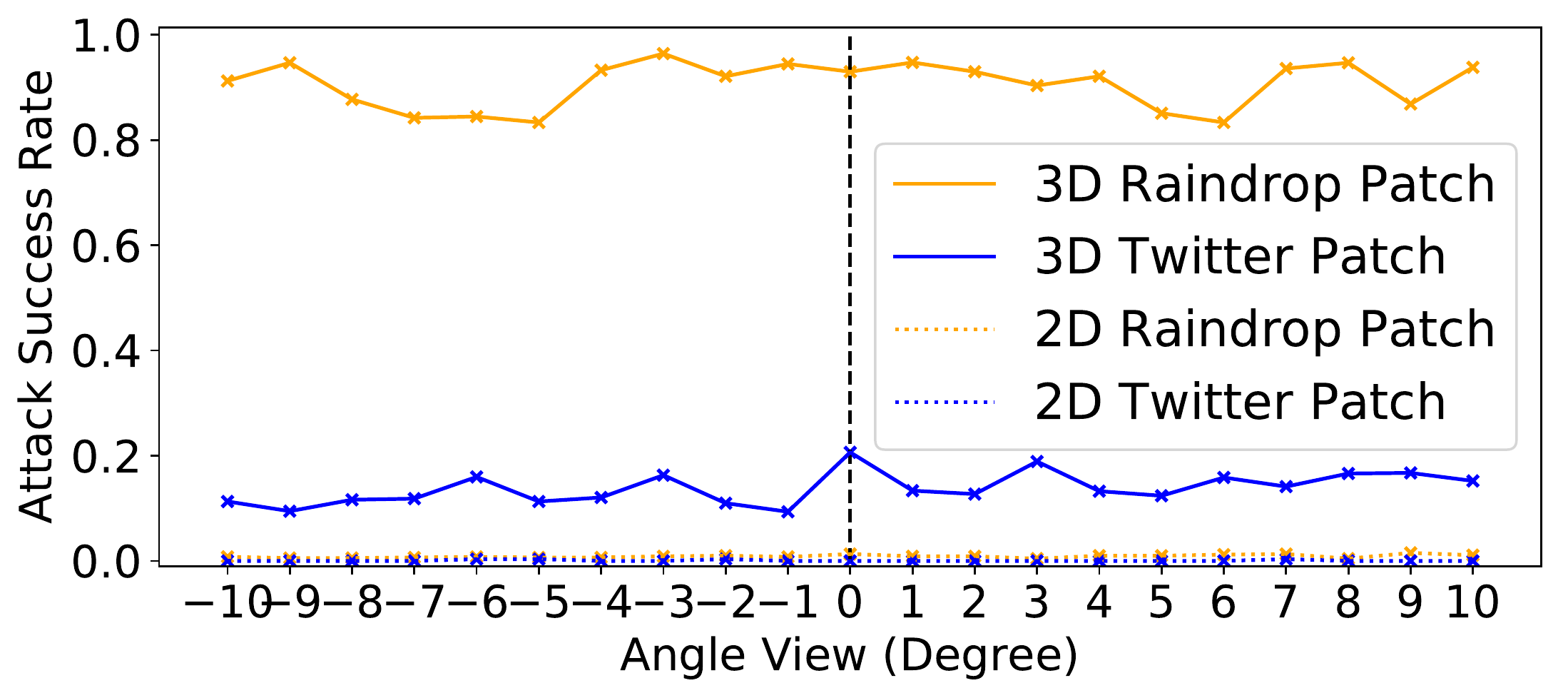}
\caption{Attack performance of two additional shapes. The 3D adversarial logo is trained in single-view (marked in dashed line) and test under multi-view attacks ($[-10,10]$ degree). We also report results of 2D patch adversarial attack baseline.}
\label{fig:extrashape}
\end{figure}

We present the results of two additional complex logo shapes to justify the generalizability of our work. The target logo shapes are \textit{Raindrop} and \textit{Twitter}, as shown in Figure \ref{fig:extralogo}. Single-view training is performed, and the performance is tested under 21 views. Figure \ref{fig:extrashape} indicates our work has the potential to tackle more intricate logo shapes, compared to 2D patch adversarial attacks. The baseline method is shown as the dashed line in Figure \ref{fig:extrashape} is the same as in Section 4.3 and the result consistently strengths our claim that the 3D adversarial logo outperforms 2D adversarial logo when 3D physical perturbations are considered.

% \clearpage
% {\small
% \bibliographystyle{ieee_fullname}
% \bibliography{egbib}
% }